\documentclass[conference]{IEEEtran}
\IEEEoverridecommandlockouts

\usepackage{cite}
\usepackage{amsmath,amssymb,amsfonts}
\usepackage[linesnumbered,ruled]{algorithm2e}
\usepackage{float}
\usepackage{booktabs}
\usepackage{multirow}
\usepackage{graphicx}
\usepackage{textcomp}
\usepackage[table]{xcolor}
\usepackage{hyperref}
\usepackage{threeparttable}
\hypersetup{
  colorlinks=true,
  linkcolor=blue,
  filecolor=blue,
  urlcolor=blue,
  citecolor=blue,
}
\usepackage[misc]{ifsym}
\usepackage{placeins}

\def\BibTeX{{\rm B\kern-.05em{\sc i\kern-.025em b}\kern-.08em
T\kern-.1667em\lower.7ex\hbox{E}\kern-.125emX}}
\begin{document}

\title{CONCORD: Asynchronous Sparse Aggregation for Device-Cloud RAG under Document Isolation}
\author{
\thanks{This work was supported in part by the National Natural Science Foundation of China (NSFC) under Grant 62302048, Grant U25A20436, and Grant 62272050; in part by Guangdong Higher Education Association under Grant 24GQN97; in part by the Guangdong Provincial Higher Education Institutions under Grant 2024KTSCX219; and in part by Beijing Normal University at Zhuhai Education Reform Project under Grant jx2025037.} \thanks{\textit{(Corresponding author: Zhiqing Tang.)}}
\IEEEauthorblockN{
Xuedong Hu$^{1,2}$,
Zhiqing Tang$^{2,1,\textrm{\Letter}}$,
Zhi Yao$^{3,2}$,
Tian Wang$^{2,5,6}$,
and Weijia Jia$^{2,4}$
}
\IEEEauthorblockA{
$^{1}${Faculty of Arts and Sciences, Beijing Normal University, Zhuhai, China}
}
\IEEEauthorblockA{
$^{2}${Institute of Artificial Intelligence and Future Networks}, {Beijing Normal University}, Zhuhai, China}
\IEEEauthorblockA{
$^{3}${School of Artificial Intelligence}, {Beijing Normal University}, Beijing, China
}
\IEEEauthorblockA{
$^{4}${Guangdong Key Lab of AI and Multi-Modal Data Processing}, {Beijing Normal-Hong Kong Baptist University}, Zhuhai, China
}
\IEEEauthorblockA{
$^{5}${College of Computer and Data Science}, {Fuzhou University}, Fuzhou, China
}
\IEEEauthorblockA{
$^{6}${Computer Science and Mathematics}, {Fujian University of Technology}, Fuzhou, China
}
Email: \{xuedonghu, yaozhi\}@mail.bnu.edu.cn,
\{zhiqingtang, tianwang, jiawj\}@bnu.edu.cn
}
\maketitle

\begin{abstract}
Retrieval-augmented generation (RAG) has emerged as a pivotal technique for improving language models by incorporating external knowledge at inference time. As device-cloud collaborative inference makes it feasible to deploy small language models on edge devices, a new setting arises in which private documents remain on the device and public knowledge resides in the cloud. Privacy and policy constraints often forbid raw document exchange, creating a document-isolated dual-end RAG setting. However, existing methods rely on frequent remote synchronization and dense evidence transfer, limiting throughput under realistic latency and bandwidth conditions. To address this issue, we propose CONCORD, an asynchronous sparse aggregation framework for dual-end RAG under document isolation. CONCORD treats the cloud as an asynchronously arriving evidence source rather than a continuously synchronized co-generator. Specifically, we introduce waiting debt control to decide whether each decoding step should continue waiting for remote participation based on the observed return of waiting. We also design a certificate-guided minimal supplementation mechanism that requests only the remote evidence needed to determine the current greedy decision. Steps that consult the cloud preserve the same greedy token as dense dual-end aggregation, while the remaining steps commit locally without remote evidence. Experiments on Natural Questions and WikiText-2 show that CONCORD improves end-to-end throughput over baselines by $1.66\times$ and $2.15\times$, respectively, while reducing per-token communication by over two orders of magnitude and maintaining comparable answer quality and perplexity.
\end{abstract}
 
 
\section{Introduction}
Retrieval-augmented generation (RAG) has become an effective way to improve compact language models by incorporating external evidence at inference time rather than relying only on parametric memory \cite{lewis2020rag,guu2020realm,khandelwal2019knnlm,izacard2021fid,borgeaud2021retro,shi2024replug,asai2024selfrag,izacard2022atlas}. In parallel, recent advances in device-cloud collaborative inference have made it increasingly feasible to deploy small language models (SLMs) on edge devices while selectively relying on the cloud for additional computation or knowledge access \cite{kang2017neurosurgeon,li2019edgeintelligence,li2018edgent,eshratifar2021jointdnn,hu2020coedge,patel2024splitwise,jin2025cecollm,narayan2025minions,xu2025edgellm,wang2025edgeintelligence,yao2024velo}. When these two trends converge, a new question arises: how should the device and the cloud jointly organize retrieval-augmented generation when each side holds its own documents?
 
This question is particularly relevant in personal-data applications, where the device serves local and privacy-sensitive context while the cloud provides access to broader public knowledge. Both sides may hold documents relevant to the next-token decision, yet privacy, policy, or system constraints often forbid raw document exchange across the two ends \cite{liu2025dragon}, forming a document-isolated dual-end RAG setting. The central problem is therefore not only how to use retrieval, but how to coordinate retrieval and generation across the device and the cloud under document isolation.
 
Existing approaches fall into two streams. The first reduces the dual-end problem to single-side generation by transferring remote documents or representations to one side \cite{guu2020realm,lewis2020rag,izacard2021fid,borgeaud2021retro,shi2024replug,asai2024selfrag,izacard2022atlas,jin2024ragcache,ma2025blockattention}, but such transfers either rebuild prefix-dependent states or incur growing communication overhead \cite{jin2024ragcache,ma2025blockattention,liu2025dragon}. The second stream keeps both sides jointly involved and performs exact aggregation online \cite{liu2025dragon}. Both streams assume that remote evidence must be made fully available—either by transferring it to one side upfront or by exchanging dense representations at every decoding step. An approach that sparsifies remote participation over both time and communication is therefore needed. To this end, two coupled challenges must be addressed.
 
\begin{figure*}[t]
\centering
\includegraphics[width=0.83\textwidth]{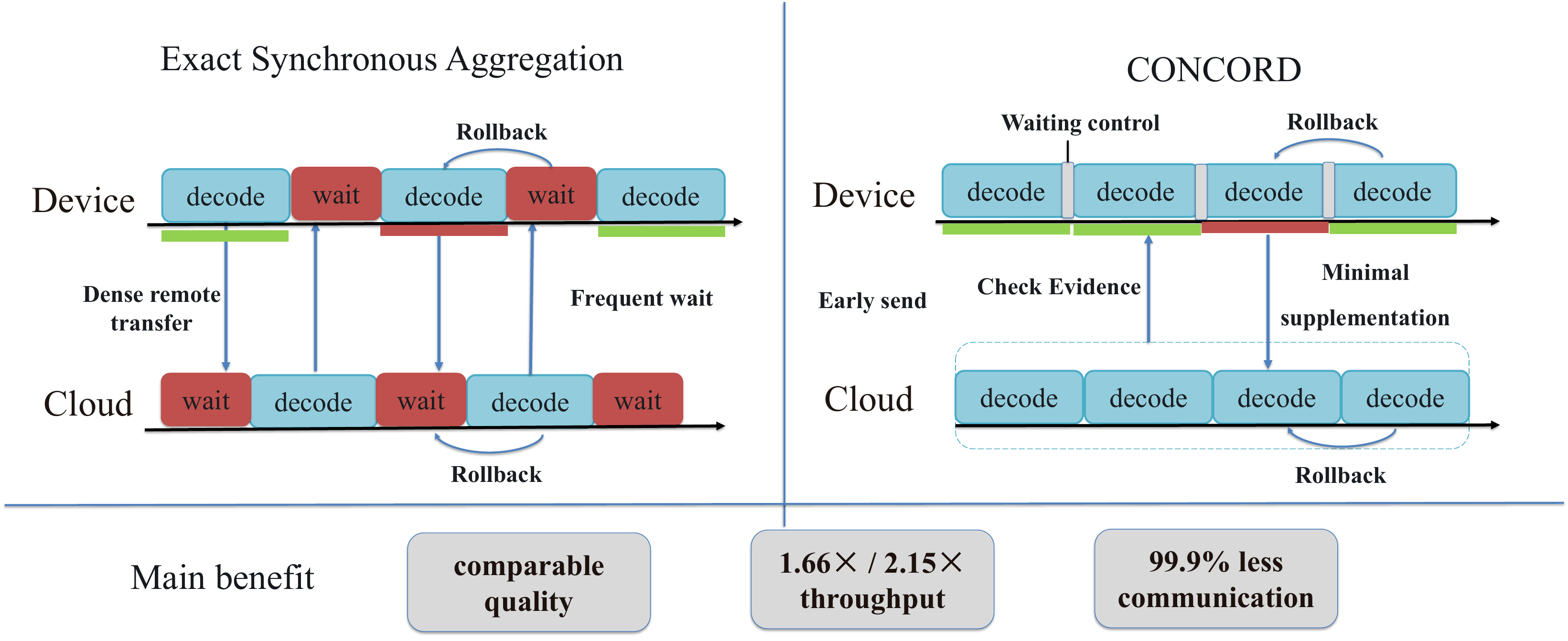}
\caption{Teaser comparison between exact synchronous aggregation and CONCORD.}
\label{fig:concord_teaser}
\end{figure*}

\textit{The first challenge is how to decide whether a decoding step should continue waiting for remote participation.} Existing dual-end methods use speculative scheduling to overlap communication with decoding, keeping the remote side on or near the critical path at nearly every step \cite{liu2025dragon}. Hiding latency through speculation alone does not remove the cost of frequent rollback and realignment when remote drafts are rejected \cite{leviathan2023specdec,chen2023specsampling}. In realistic deployments, the arrival delay of remote evidence reflects not only network conditions but also these residual rollback effects. As this delay accumulates, the local side increasingly has enough evidence to decide on its own, so continued waiting mainly adds latency without changing the output.
 
\textit{The second challenge is how to determine the minimal remote evidence needed to resolve the current greedy decision.} When a step does consult the remote side, both centralized and dual-end methods transfer full evidence representations, whether as complete KV states \cite{kang2017neurosurgeon,li2018edgent,eshratifar2021jointdnn,hu2020coedge} or dense remote score vectors \cite{liu2025dragon}. Yet language-model output distributions are typically concentrated \cite{leviathan2023specdec,chen2023specsampling}: a small number of candidates occupy most of the probability mass, and only these candidates can affect the final decision. Transferring evidence for the remaining candidates wastes bandwidth without changing the committed token.
 
These two challenges are tightly coupled: reducing waiting alone is insufficient if consultation still triggers large remote transfers, while reducing payload alone is insufficient if the remote side is still invoked too frequently. The key issue is how to jointly control when remote participation should occur and how much evidence should be revealed once consultation begins. Figure~\ref{fig:concord_teaser} illustrates this contrast.
 
In this paper, we present CONCORD, an asynchronous sparse aggregation framework for dual-end RAG under document isolation. CONCORD introduces two coupled mechanisms. First, waiting debt control uses the observed return of waiting to decide whether the current decoding step should continue waiting for remote participation, and shortens waiting when further waiting is unlikely to pay off. Second, certificate-guided minimal supplementation requests only the token-level remote evidence that may still change the current greedy decision, and stops as soon as sufficient evidence is available. In this way, CONCORD treats the cloud not as a continuously synchronized co-generator, but as an asynchronously arriving source of remote evidence. Unlike speculative scheduling, which keeps the remote side on the critical path and hides latency through parallelism, CONCORD removes the remote side from the critical path entirely on steps where local evidence suffices. Under greedy decoding, once a step enters the supplementation pipeline, certificate success or exact fallback returns the same greedy token as dense consulted-step aggregation, so the communication-layer exactness target is preserved on consulted steps. Together, these two components decouple the temporal decision of whether remote participation should occur from the communication decision of how much remote evidence should be revealed once consultation begins.
 
The main contributions are summarized as follows.
\begin{enumerate}
\setlength{\itemsep}{0pt}
\setlength{\topsep}{2pt}
\setlength{\parsep}{0pt}
\setlength{\partopsep}{0pt}
\item \textbf{Dual-end sparse aggregation formulation and CONCORD framework.} We formulate dual-end RAG under document isolation as jointly minimizing generation loss and device-cloud interaction cost along two dimensions: the temporal frequency of remote participation and the per-consultation communication payload. Based on this formulation, we propose CONCORD, an asynchronous sparse aggregation framework that reformulates the cloud as an asynchronously arriving evidence source rather than a continuously synchronized co-generator.
\item \textbf{Waiting debt control and certificate-guided minimal supplementation.} We design two coupled mechanisms to address the two challenges identified above. Waiting debt control tracks the observed return of remote participation through a lightweight debt queue and adaptively shortens waiting when further blocking is unlikely to change the output, with only $O(1)$ additional computation per token. Certificate-guided minimal supplementation uses an upper-bound test to certify the current greedy decision early, and requests only the ambiguity-critical token ids when certification fails, preserving the dense consulted-step greedy result through certificate success or exact fallback.
\item \textbf{Evaluation on Natural Questions and WikiText-2.} Experiments including both controlled single-machine and real two-machine deployments show that CONCORD improves end-to-end throughput over DRAGON by about $1.66\times$ and $2.15\times$, respectively, while reducing per-token communication by over two orders of magnitude on both tasks and maintaining comparable answer quality and perplexity.
\end{enumerate}
 
The remainder of this paper is organized as follows. Section~\ref{sec:related} reviews related work. Section~\ref{sec:preliminaries} introduces the preliminaries and problem formulation. Section~\ref{sec:method} presents the methodology of CONCORD. Section~\ref{sec:experiments} reports experimental results. Finally, Section~\ref{sec:conclusion} concludes the paper.
 
\section{Related Works}\label{sec:related}
\textbf{RAG with external memory and multiple documents.}
Early retrieval-augmented language models such as REALM and kNN-LM attached non-parametric memory or nearest-neighbor retrieval to language-model inference and training \cite{guu2020realm,khandelwal2019knnlm}. Subsequent work, including RAG, FiD, RETRO, REPLUG, SELF-RAG, and Atlas, strengthened multi-document reasoning through retrieval-conditioned decoding or training \cite{lewis2020rag,izacard2021fid,borgeaud2021retro,shi2024replug,asai2024selfrag,izacard2022atlas}. RAGCache and Block-Attention further reduced the serving overhead of long retrieved context through cache reuse and more efficient prefilling \cite{jin2024ragcache,ma2025blockattention}. All these methods assume that retrieved evidence can be centralized into a single decoding context, either by merging documents directly or by transferring cached states. Under document isolation, however, raw cross-end document exchange is forbidden, so the aggregation must happen in the output space rather than in the input context.
 
\textbf{Edge-cloud collaborative inference and service systems.}
Collaborative intelligence systems have explored how to split DNN execution across the device, edge, and cloud for latency, energy, or privacy goals \cite{kang2017neurosurgeon,li2019edgeintelligence,li2018edgent,eshratifar2021jointdnn,hu2020coedge,huang2022lightweight}. Service-level work further studied edge-cloud coordination under QoS and resource constraints \cite{sun2020iotconfig,xiang2021energy,nkenyereye2023drl}. More recently, CE-CoLLM, CEED, C$^2$F, EdgeLLM, and VELO have revisited collaborative inference for large or foundation models, addressing partitioning, context-aware offloading, QoS optimization, or model-side acceleration \cite{jin2025cecollm,chen2025ceed,zhao2025c2f,feng2025distributedllm,xu2025edgellm,yao2024velo,yao2025llmqos,li2025cloudedge,mou2026adaptive}. DRAGON extended this line to distributed RAG, formulating exact output aggregation across device-side and cloud-side corpora \cite{liu2025dragon}. None of these methods, however, addresses the question of when remote participation is worth its cost or how much remote evidence is actually needed per consultation. They either assume continuous joint execution or focus on one-time partitioning decisions, leaving the sparse-participation regime unexplored.
 
\textbf{Speculative decoding and LLM serving.}
Speculative decoding accelerates autoregressive generation by letting a draft process run ahead of a verifier and then accepting or correcting the draft \cite{leviathan2023specdec,chen2023specsampling,cai2024medusa,bhendawade2024specstream}. Recent variants such as recurrent drafters, EAGLE, and AdaSpec further improved drafting efficiency or SLO-aware serving \cite{zhang2024drafter,li2024eagle,huang2025adaspec}. Serving systems such as DistServe and Sarathi-Serve optimized prefill-decode scheduling and throughput-latency tradeoffs for centralized LLM stacks \cite{zhong2024distserve,agrawal2024sarathi}. These ideas informed our use of rollback, preemption, and asynchronous execution, but they target single-model or centralized pipelines and do not address the dual-end question of when to consult the remote side and how much evidence to transfer per consultation.
 
In summary, existing RAG methods assume centralized evidence access, edge-cloud collaborative systems assume continuous joint execution, and speculative decoding targets single-model pipelines. CONCORD bridges these lines by introducing sparse, asynchronous remote participation for exact dual-end aggregation under document isolation.

\section{Preliminaries}\label{sec:preliminaries}
\subsection{Retrieval-Augmented Generation}
In retrieval-augmented generation (RAG), a language model incorporates external documents retrieved from a corpus at inference time rather than relying only on parametric memory \cite{guu2020realm,lewis2020rag,izacard2021fid,borgeaud2021retro,shi2024replug,asai2024selfrag,izacard2022atlas}. Given the current prefix $x_{<t}$ and a retrieved document set $D$, each document $d \in D$ induces a document-conditioned next-token distribution $p(x_t \mid d, x_{<t})$. Following the output-aggregation view of multi-document RAG \cite{lewis2020rag,shi2024replug}, the target next-token distribution is
\begin{equation}
p(x_t \mid x_{<t})
=
\sum_{d \in D} \omega_t(d)\, p(x_t \mid d, x_{<t}),
\label{eq:rag_agg}
\end{equation}
where $\omega_t(d)$ is the relevance weight of document $d$ at step $t$. In practice, the summation runs over a top-$k$ retrieved subset. Equation~(\ref{eq:rag_agg}) shows that the aggregation operates on document-conditioned output distributions, or equivalently on their vocabulary-level scores. This view suits distributed settings because it separates document-local inference from the final decision over the shared vocabulary.
 
\subsection{Device-Cloud Distributed RAG}
To support personal-data applications, we consider a dual-end setting in which retrieved evidence is naturally split across the device and the cloud. Let $D^{\text{device}}$ denote the device-side private document set and $D^{\text{cloud}}$ denote the cloud-side public document set, with
\begin{equation*}
D = D^{\text{device}} \cup D^{\text{cloud}}.
\end{equation*}
Each side runs its own retrieval and document-conditioned inference without exchanging raw documents. At decoding step $t$, the device and cloud produce vocabulary-level scores from $D^{\text{device}}$ and $D^{\text{cloud}}$, respectively. The final distribution is obtained by aggregating contributions from both sides in the output space rather than by merging documents into one centralized context.
 
This workflow is a distributed version of Equation~(\ref{eq:rag_agg}). Working in the log domain, we write the device-side and cloud-side log-scores as $\ell_L^t(v)$ and $\ell_R^t(v)$ for each token $v \in V$. Let $\pi_L^t=\sum_{d\in D^{\text{device}}}\omega_t(d)$ and $\pi_R^t=\sum_{d\in D^{\text{cloud}}}\omega_t(d)$ be the per-side relevance weights. The aggregated log-score $\ell_T^t(v)$ is then obtained by combining both sides through a weighted log-sum-exp operation; the exact formula is given in Section~\ref{sec:comm}. Greedy decoding returns
\begin{equation*}
y_t = \arg\max_{v \in V} \ell_T^t(v).
\end{equation*}
Unlike centralized RAG, the two sides cannot exchange documents and rerun a single-model pipeline. How to coordinate this process efficiently under document isolation is the problem we formulate next.
 
\subsection{Problem Formulation}
The goal of dual-end aggregation is to preserve decision quality while reducing the system cost induced by device-cloud interaction. Let $F=(F^{\text{device}}, F^{\text{cloud}})$ denote the joint generation strategy of the two sides. At each decoding step, the system may exchange auxiliary information $\mathcal{M}_t$ between the two sides and then produce a distribution over the next token. The optimization objective is to jointly minimize generation loss and interaction cost:
\begin{equation*}
\min_{F}\;
\frac{1}{T}\sum_{t=1}^{T}
\Big(
\mathcal{L}_t
+\,
\lambda\, C_t(\mathcal{M}_t, F)
\Big),
\end{equation*}
where $\mathcal{L}_t$ is the task loss at step $t$, $C_t(\mathcal{M}_t, F)$ is the system cost from communication and waiting, and $\lambda$ controls the quality-efficiency tradeoff. Exact dual-end methods such as DRAGON \cite{liu2025dragon} reduce this cost mainly by hiding synchronization delay through speculative scheduling. We instead aim to reduce the frequency of remote participation and the volume of remote evidence transferred per consultation.
 
This formulation highlights two dimensions that drive the rest of our design. Along the temporal dimension, the system must decide whether the current step should continue waiting for remote participation. Along the communication dimension, it must determine how much remote evidence to reveal before the greedy decision is fixed. CONCORD addresses both dimensions jointly while preserving the output-aggregation target defined above for consulted steps.
 
\section{Methodology}\label{sec:method}
\subsection{Framework Overview}
\begin{figure}[t]
\centering
\includegraphics[width=\columnwidth,trim=6 6 6 6,clip]{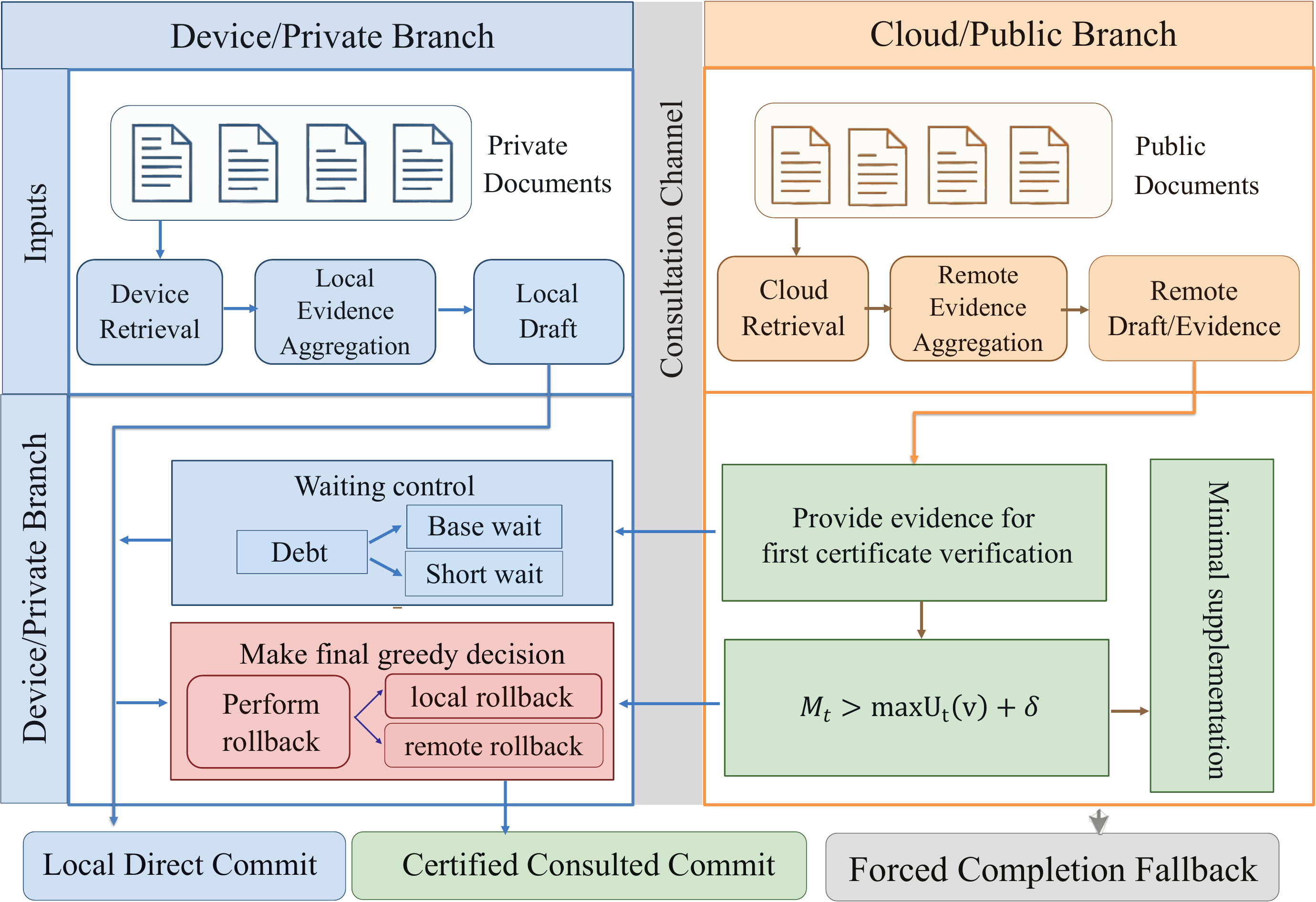}
\caption{Overview of CONCORD.}
\label{fig:concord_overview}
\end{figure}

As illustrated in Figure~\ref{fig:concord_overview}, CONCORD consists of a local decision branch and a remote evidence branch. The device continues decoding with private evidence, while the cloud prepares public evidence asynchronously. Following the optimization objective in Section~\ref{sec:preliminaries}, CONCORD reduces end-to-end system cost along both the temporal and communication dimensions while preserving the output-aggregation target $y_t = \arg\max_{v \in V} \ell_T^t(v)$ on steps that enter remote consultation. It does so through two coupled mechanisms: waiting control and communication control.

\subsection{Waiting Control}
The design of waiting control is motivated by an empirical observation that we verify in Section~\ref{sec:experiments}. In most decoding steps, the local decision already matches the dense dual-end result, and only a small number of positions remain sensitive to remote evidence. The cloud therefore serves as a \emph{proofreading} source rather than a continuously synchronized co-generator. Instead of aggregating remote evidence at every step, the remote side intervenes only where public evidence may still overturn the local decision.
 
We therefore replace dual-end synchronous progression with asynchronous progression. The device keeps generating with private evidence, while the cloud prepares proofreading evidence in the background and sends it whenever ready. If the received evidence is sufficient to confirm the current local decision, the device keeps the local token and moves on; the concrete certificate rule is described in Section~\ref{sec:comm}. If the local and remote drafts disagree, the system evaluates the aggregated distribution before committing. The losing side then rolls back and realigns to the committed token.
 
In this asynchronous framework, the main throughput bottleneck is rollback-induced waiting. Each rollback is not only a cost for the current step; it also propagates delay to later remote arrivals because the losing side must realign before preparing useful evidence again. Frequent rejection of local drafts indicates that the remote side is providing useful corrections, so preparing evidence ahead of time is worthwhile. Conversely, frequent rejection of remote drafts means the device waits for evidence that is ultimately discarded. Rejection behavior is therefore a direct signal of the value of waiting. We design a waiting controller that adapts the waiting budget according to the observed long-run return of remote participation.
 
Let $T_t^L$ be the time when the local side enters the decision at step $t$, and let $T_t^R$ be the time when the current remote evidence becomes consumable. The remote arrival delay is
\[
D_t = T_t^R - T_t^L.
\]
This delay reflects both the current network and computation gap and the residual effect of earlier rollback. When $D_t$ exceeds the current waiting budget $\tau_t$, the step times out. As timeouts accumulate, the local side is increasingly likely to have enough evidence to decide on its own, so the marginal value of waiting declines. To suppress long-run low-return waiting, we introduce a waiting debt queue
\[
q_{t+1}
=
\min\Big(
Q_{\max},
\big[q_t+\xi_t-\rho_a \alpha_t\big]_+
\Big),
\]
where $\xi_t\in\{0,1\}$ indicates that the current step times out or fails to obtain useful remote evidence, $\alpha_t\in\{0,1\}$ indicates that remote evidence is effectively absorbed, and $\rho_a>0$ controls the repayment rate. Timeout or unproductive waiting increases debt, while effective proofreading repays it. The system then selects one of two waiting budgets:
\[
\tau_t=
\begin{cases}
\min\{\tau_{\mathrm{base}},\tau_{\mathrm{short}}(q_t)\}, & q_t\ge Q,\\
\tau_{\mathrm{base}}, & q_t<Q.
\end{cases}
\]
Algorithm~\ref{alg:waiting_control} summarizes the resulting discrete rule.
 
\begin{algorithm}[t]
\caption{Waiting Debt Control}
\label{alg:waiting_control}
\textbf{Input:} debt $q_t$, waiting budgets $\tau_{\mathrm{base}}$ and $\tau_{\mathrm{short}}(\cdot)$, threshold $Q$, debt cap $Q_{\max}$, repayment factor $\rho_a$, local and remote drafts\;
\textbf{Output:} committed token and updated debt $q_{t+1}$\;
\eIf{$q_t \ge Q$}{
  $\tau_t \gets \min\{\tau_{\mathrm{base}}, \tau_{\mathrm{short}}(q_t)\}$\;
}{
  $\tau_t \gets \tau_{\mathrm{base}}$\;
}
Wait up to $\tau_t$ for the remote draft\;
\If{timeout}{
  Commit the local token without remote certification\;
  $q_{t+1} \gets \min(Q_{\max}, q_t + 1)$\;
  Clear remote cache and advance remote epoch\;
  \KwRet\;
}
Aggregate local and remote draft\;
Obtain $y_t$ and $\alpha_t$\;
\eIf{$\alpha_t = 1$}{
  $q_{t+1} \gets \max(0, q_t - \rho_a)$\;
}{
  Submit the local token and realign remote state\;
  $q_{t+1} \gets \min(Q_{\max}, q_t + 1)$\;
  Clear remote cache and advance remote epoch\;
}
\end{algorithm}
 
When debt is low, the system uses the base waiting budget and gives the cloud normal opportunities to participate. Once debt crosses the threshold $Q$, the system shortens waiting and lets the device move forward earlier. If remote evidence later becomes useful again, debt is repaid automatically and the controller relaxes back toward the base regime.
 
\subsection{Communication Control}\label{sec:comm}
Language-model output distributions are typically concentrated on a small number of candidates that occupy most of the probability mass. Full remote transfer is therefore usually unnecessary; only the candidates competitive enough to overturn the current greedy decision matter. We design the communication layer around this residual ambiguity and apply progressive supplementation guided by an upper-bound certificate, transmitting only the ambiguity-relevant remote evidence while preserving the dense consulted-step greedy result.
 
For any consulted step, let $\Omega_t=V$ denote the complete remote reference representation and let
\[
y_t^{\mathrm{dense}}
=
\arg\max_{v\in\Omega_t}\ell_T^t(v)
\]
be the greedy token under dense aggregation. The local side first receives a known subset $K_t\subseteq\Omega_t$ with exact cloud scores $\hat \ell_R^t(v)$ for $v\in K_t$ and a tail bound $b_t$ on the unrevealed remainder. Using the per-side weights $\pi_L^t$ and $\pi_R^t$ defined in Section~\ref{sec:preliminaries} and writing $\operatorname{LSE}(a,b)=\log(e^a+e^b)$, the exact aggregated score for every revealed candidate $v\in K_t$ is
\[
\ell_T^t(v)
=
\operatorname{LSE}\!\big(\log \pi_L^t+\ell_L^t(v),\ \log \pi_R^t+\hat \ell_R^t(v)\big).
\]
Let the current revealed maximum be
\[
M_t=\max_{v\in K_t}\ell_T^t(v).
\]
For any unrevealed candidate $v\in\Omega_t\setminus K_t$, the most optimistic score it can still attain under the current tail bound is
\[
U_t(v)=
\operatorname{LSE}\!\big(\log \pi_L^t+\ell_L^t(v),\ \log \pi_R^t+b_t\big).
\]
If
\[
M_t > \max_{v\in\Omega_t\setminus K_t} U_t(v) + \delta,
\]
where $\delta\ge 0$ is a certification tolerance, then the current incumbent is already certified and
\[
\arg\max_{v\in K_t}\ell_T^t(v)=y_t^{\mathrm{dense}}.
\]
Otherwise we collect the still-competitive candidates in the ambiguity set
\[
A_t=
\left\{
v\in \Omega_t\setminus K_t:\ U_t(v)\ge M_t-\delta
\right\}.
\]
We then rank $A_t$ by normalized local mass
\[
\tilde p_t(v)=
\frac{\exp(\ell_L^t(v))}
{\sum_{u\in A_t}\exp(\ell_L^t(u))},
\qquad v\in A_t,
\]
and choose the minimal supplementation set
\[
\begin{gathered}
S_t^\star
=
\arg\min_{S\subseteq A_t}|S|\\
\text{s.t.}\quad
\sum_{v\in S}\tilde p_t(v)\ge 1-\epsilon,
\end{gathered}
\]
where $\epsilon\in(0,1)$ is a coverage threshold that controls how much local mass must be covered before supplementation stops.
 
Each consulted step therefore proceeds through up to four stages: certificate check, token-id query, sparse-chunk supplementation, and forced completion. Algorithm~\ref{alg:progressive_supplementation} summarizes the procedure.
\begin{algorithm}[t]
\caption{Certificate-Guided Supplementation}
\label{alg:progressive_supplementation}
\textbf{Input:} partial state $(K_t,b_t)$, candidate set $\Omega_t$, local scores $\ell_L^t(\cdot)$, weights $(\pi_L^t,\pi_R^t)$, query budget $k$, round budget $R_{\max}$\;
\textbf{Output:} exact greedy token\;
$r \gets 0$\;
\While{$r < R_{\max}$}{
  Use $\ell_L^t(\cdot)$ and $(\pi_L^t,\pi_R^t)$ to compute $M_t$ on $K_t$ and upper bounds $U_t(v)$\;
  \If{the certificate is valid}{
    \KwRet $\arg\max_{v \in K_t}\ell_T^t(v)$\;
  }
  $A_t \gets \{v \in \Omega_t \setminus K_t : U_t(v) \ge M_t - \delta\}$\;
  Rank $A_t$ by $\tilde p_t(v) \propto \exp(\ell_L^t(v))$\;
  \eIf{query enabled and $A_t \neq \emptyset$}{
    Request up to $k$ top-ranked token ids from $A_t$ and merge replies into $K_t$\;
  }{
    Request the next sparse chunk and update $(K_t, b_t)$\;
  }
  $r \gets r + 1$\;
}
Force completion of the remaining $\Omega_t$\;
\KwRet exact $\arg\max_{v \in \Omega_t}\ell_T^t(v)$\;
\end{algorithm}
The communication layer therefore requests only the remote evidence necessary to fix the current decision. Candidates outside $A_t$ are too weak to change the outcome, so their exact remote scores are never transmitted. If the certificate holds, the device commits immediately; otherwise, it queries a few ambiguity-critical token ids first and resorts to larger sparse chunks only when needed. As a result, the transmitted payload scales with the residual ambiguity rather than with the full vocabulary size.
 
\subsection{Unified View}
Waiting control and communication control can be viewed under a single lens. The waiting budget $\tau_t$ governs whether a step keeps the remote side on the critical path, while the supplementation set $S_t^\star$ governs how much evidence is transmitted once consultation begins. This yields the following unified objective:
\[
\begin{aligned}
\min_{\{\tau_t,S_t^\star\}}\quad
&\frac{1}{T}\sum_{t=1}^{T}\mathbb{E}\!\left[
\lambda_1\,\mathrm{Latency}_t(\tau_t)
+\lambda_2\,|S_t^\star|
+\lambda_3\,\mathrm{Fallback}_t
\right]\\
\text{s.t.}\quad
&\frac{1}{T}\sum_{t=1}^{T}\mathbb{E}\big[\xi_t-\rho_a \alpha_t\big]\le 0.
\end{aligned}
\]
The objective balances three costs: waiting latency, the size of remote evidence transmitted per consultation, and the cost of fallback to full completion when certification fails. The constraint prevents long-run accumulation of unproductive waiting by bounding the expected net debt growth.
 
\subsection{Theoretical Analysis}
\textbf{Time and Communication Complexity.}
Dense consultation is a special case of CONCORD in which every step keeps the remote side on the critical path and communication control always falls through to forced completion. CONCORD therefore does not incur a worse worst-case cost than the dense baseline. If waiting remains worthwhile and no certificate can terminate early, the procedure degenerates to full consultation.
 
Along the time dimension, the waiting controller adds only constant overhead per token. At each step it reads $q_t$, compares it with $Q$, selects $\tau_t$, and updates $q_{t+1}$ from the binary outcomes $(\xi_t,\alpha_t)$, so its computation and state complexity are both $O(1)$. When remote evidence is repeatedly useful, fewer rollbacks amortize the waiting cost; when remote participation is repeatedly rejected, the controller shortens waiting and shifts the system toward local commitment.
 
Along the token dimension, let $|\Omega_t|$ be the size of the full remote reference representation at step $t$, $|K_t|$ the currently known subset, and $|S_t^\star|$ the minimal supplementation size returned by the ambiguity rule. In the worst case, repeated certificate failure triggers forced completion, so the transmitted evidence remains $O(|\Omega_t|)$, matching dense consultation. In the typical sparse case, only ambiguity-critical candidates need to be transmitted. The effective communication cost per consulted step then becomes $O(|K_t^{(0)}|+|S_t^\star|)$, where $|K_t^{(0)}|$ is the initial known-set size and $|S_t^\star|$ is the cumulative supplementation across all rounds. When $|S_t^\star| \ll |\Omega_t|$, the cost is governed by unresolved ambiguity rather than by the full remote score vector.
 
Taken together, CONCORD preserves the dense consulted-step cost bound in the worst case while reducing average waiting and transmission whenever disagreement is sparse over time and probability mass is concentrated on a few competitive candidates. The gain is largest when only a small fraction of steps require remote correction and, within those steps, only a few token ids can overturn the greedy decision.
 
\section{Experiments}\label{sec:experiments}
\subsection{Setup}
\textbf{Testbed.}
We evaluate CONCORD in a three-process dual-end deployment consisting of a device-side generation process, a cloud-side generation process, and a retrieval service. All controlled experiments first run on a single machine to isolate the effect of remote-participation control. The CPU is an Intel Xeon Platinum 8352V (dual socket, 72 physical cores / 144 threads), the memory is 125\,GB, and the GPU setup is 4$\times$ RTX 4090D (24\,GB). The device-side generation process runs on \texttt{cuda:3}, the cloud-side generation process on \texttt{cuda:1}, and the retrieval service on \texttt{cuda:0}. The software stack is Python 3.10, PyTorch 2.5.1, and Transformers 4.57.3. Both ends communicate through TCP sockets with latency injected directly into the transport path for reproducibility. Unless otherwise specified, RTT is set to 50\,ms.
 
\textbf{Datasets.}
We evaluate on two tasks. For answer generation, we use the Natural Questions (NQ) development set, randomly sampling 500 examples and repeating evaluation with 3 random seeds; each example generates 10 new tokens. For long-form continuation, we use the WikiText-2 test set, sampling 100 examples with the first 64 tokens as the prefix and the following 48 tokens as the target. Under teacher-forcing evaluation with a fixed model, per-example PPL variance is low, so this subset size suffices to distinguish methods that share the same aggregation target (Table~\ref{tab:wikitext_quality}). Mechanism-dissection experiments follow the same NQ pipeline but use 100 examples with 3 random seeds to keep repeated instrumentation affordable.
 
\textbf{Metrics.}
For NQ, we report F1 and EM as quality metrics. For WikiText-2 long-form continuation, we report teacher-forcing perplexity (PPL) on the same examples used in the system experiments. For system efficiency, NQ reports end-to-end queries per second (QPS), while WikiText-2 reports generation throughput in tok/s. Communication is measured uniformly as transmitted and received bytes per generated token.
 
\textbf{Models and baselines.}
We use Qwen3-1.7B as the generation model on both ends. Retrieval is always dual-end and sharded: the device side and the cloud side connect to different local retrieval instances, so they access non-overlapping evidence shards. Each query retrieves four documents in total, with two retained on the device side and two on the cloud side. We evaluate the following methods throughout the paper:
\begin{itemize}
\setlength{\itemsep}{0pt}
\setlength{\topsep}{0pt}
\setlength{\parsep}{0pt}
\setlength{\partopsep}{0pt}
  \item \textbf{CRCG}: centralized generation augmented with retrieval from the device-side corpus only, using the context-aggregation strategy.
  \item \textbf{DRCG}: on-device generation augmented with evidence retrieved from the distributed corpus spanning both the device and the cloud, using the context-aggregation strategy.
  \item \textbf{DRDG/TW}: distributed retrieval-augmented generation using the output-aggregation strategy with token-wise synchronization during decoding.
  \item \textbf{DRDG/SW}: distributed retrieval-augmented generation using the output-aggregation strategy with sequence-wise synchronization, i.e., one-time aggregation of independently generated sequences from the device and the cloud.
  \item \textbf{DRAGON}: a speculative dual-end output-aggregation method that reduces token-wise synchronization waiting through parallel generation and scheduling.
  \item \textbf{CONCORD}: our method, which reduces remote participation through waiting debt control and certificate-guided supplementation.
  \item \textbf{CONCORD-NoSystem}: CONCORD with waiting control removed while keeping the supplementation pipeline.
  \item \textbf{CONCORD-NoComm}: CONCORD with the communication design removed while keeping waiting control.
\end{itemize}
Among these methods, DRAGON is the only existing approach that performs exact output-space aggregation under document isolation. Other edge-cloud methods such as CE-CoLLM \cite{jin2025cecollm} and EdgeLLM \cite{xu2025edgellm} target model partitioning or offloading without cross-end retrieval aggregation, so they are not directly comparable. DRDG/TW and DRDG/SW already represent two extreme synchronization strategies (per-token and per-sequence), while DRAGON adds speculative scheduling. The ablation variants NoSys and NoComm further serve as simpler alternatives: NoSys applies sparse communication without adaptive waiting, and NoComm applies adaptive waiting without sparse communication.

\textbf{Implementation details.}
Within each end, retrieved evidence is first merged into a single intra-end representation before inter-end aggregation, so the comparison isolates coordination overhead rather than intra-end fusion details. On the communication path, CONCORD transmits compact remote evidence packets instead of full dense score vectors. A fixed-length message header is sent over TCP sockets, payloads are serialized as binary tensors and compressed before transmission. When a step remains uncertified, the communication layer first queries ambiguity-critical token ids and resorts to sparse chunks only when necessary. The cloud retains the current-step logit vector in GPU memory and responds to each token-ID query by indexing the requested entries, so the per-query overhead is a GPU gather plus one TCP round trip. The tail bound $b_t$ is initialized to the global maximum of the remote log-score vector and is tightened after each sparse-chunk reply to the maximum over unrevealed positions. On the decoding side, generation is preemptible. Each model layer checks for a stop event so that rejected drafts can be interrupted early; the caller then rolls back KV-cache state and resumes from the corrected token.
\subsection{Main Results}
\begin{figure}[!t]
\centering
\includegraphics[width=0.8\columnwidth]{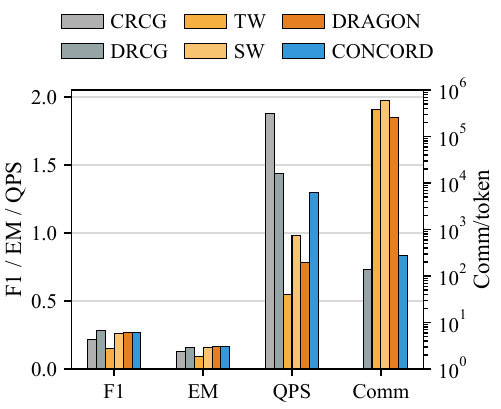}
\caption{Main results on Natural Questions. All six methods are compared on answer quality (F1, EM), throughput (QPS), and communication cost (bytes/token).}
\label{fig:nq_main_results}
\end{figure}

\textbf{Performance.}
We first examine whether CONCORD preserves task quality while reducing the system path. Figure~\ref{fig:nq_main_results} summarizes the main NQ results. CONCORD remains matched with DRAGON in answer quality: F1 is $0.267\pm0.006$ vs.\ $0.268\pm0.008$, and EM is $0.167\pm0.005$ vs.\ $0.169\pm0.005$ (CONCORD vs.\ DRAGON). The gap stays within across-seed variation. Among the other output-aggregation baselines, DRDG/TW shares the same token-wise exact aggregation target and therefore achieves comparable quality, whereas DRDG/SW aggregates two independently generated sequences only once, trading per-token exactness for lower synchronization frequency. Table~\ref{tab:wikitext_quality} shows the same pattern on WikiText-2. Methods that share the same output-aggregation form (DRDG/TW, DRAGON, CONCORD) all achieve 17.016 PPL, while CRCG and DRCG yield higher perplexity because they use different evidence organizations.
 
\begin{table}[!t]
\caption{Quality results on WikiText-2.}
\label{tab:wikitext_quality}
\centering
\small
\begin{tabular}{@{}p{0.18\columnwidth}p{0.50\columnwidth}r@{}}
\toprule
Method & Evidence form & PPL$\downarrow$ \\
\midrule
CRCG & Single-end local shard & 21.351 \\
DRCG & Dual-end evidence merged into single-end context & 17.179 \\
DRDG/TW & Dual-end output aggregation (token-wise sync) & 17.016 \\
DRAGON & Dual-end output aggregation & 17.016 \\
CONCORD & Dual-end output aggregation & 17.016 \\
\bottomrule
\end{tabular}
\end{table}
 
\begin{figure}[!t]
\centering
\includegraphics[width=0.8\columnwidth]{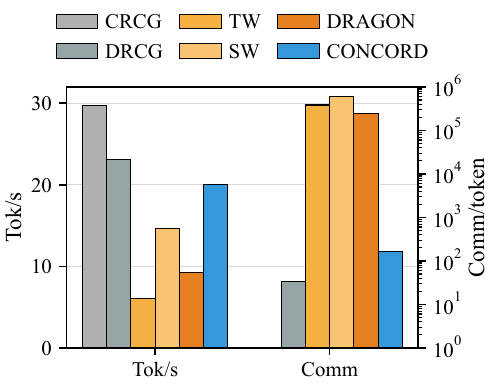}
\caption{System efficiency on WikiText-2. All six methods are compared on generation throughput (tok/s) and communication cost (bytes/token).}
\label{fig:wikitext_main_results}
\end{figure}

\textbf{Efficiency.}
The efficiency gap is much larger. On NQ (Figure~\ref{fig:nq_main_results}), CONCORD raises throughput from $0.782\pm0.006$ to $1.295\pm0.004$ QPS relative to DRAGON (about $1.66\times$), while communication drops from $253479.7\pm1748.7$ to $282.7\pm25.3$ bytes per token, a reduction of about 99.9\%. Figure~\ref{fig:wikitext_main_results} shows the same trend on WikiText-2: under the same aggregation target, throughput rises from 9.29 to 20.00 tok/s (about $2.15\times$) and per-token communication falls from 243640.3 to 160.9 bytes (again about 99.9\%). Among the other baselines, DRDG/TW transmits a full dense score vector at every token and blocks until the cloud replies, yielding the highest communication cost and the lowest throughput. DRDG/SW avoids per-token synchronization but pays for it with degraded quality. CRCG and DRCG involve no inter-end score exchange, yet they cannot match the quality of output-aggregation methods. These results confirm that the same quality target can be reached at a much lower system cost when remote participation is sparsified along both dimensions.
\subsection{Ablation Studies}
\begin{figure}[!t]
\centering
\includegraphics[width=0.8\columnwidth]{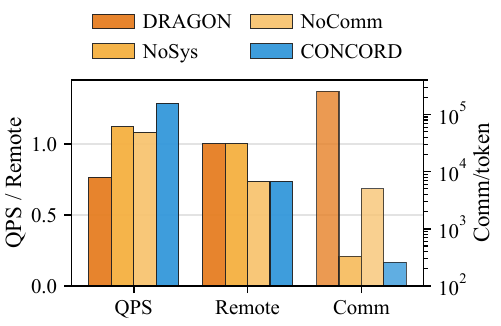}
\caption{Ablation results on Natural Questions. DRAGON, NoSys, NoComm, and CONCORD are compared on throughput (QPS), remote participation ratio, and communication cost (bytes/token).}
\label{fig:ablation_breakdown}
\end{figure}

To isolate the contribution of each component, we conduct ablation studies on 100 NQ development examples with 3 random seeds at RTT\,=\,50\,ms. The four compared variants are DRAGON, CONCORD-NoSystem (NoSys), CONCORD-NoComm (NoComm), and CONCORD (Figure~\ref{fig:ablation_breakdown}). All variants preserve answer quality (F1 and EM remain within across-seed variation), so the discussion below focuses on efficiency.
 
\textbf{Waiting control.}
Waiting control mainly affects time-domain sparsification. Removing it pushes the remote participation ratio back to 1.000, meaning the remote side again remains on the critical path at nearly every step. Throughput then falls from 1.286 to 1.123 even though communication stays low at 331.0 bytes/token. This mirrors the causal chain described in Section~\ref{sec:method}: without debt-aware waiting, the system still performs sparse communication but keeps paying unnecessary blocking cost.
 
\textbf{Communication control.}
Communication control determines how much evidence each consulted step transmits. When it is disabled, throughput decreases from 1.286 to 1.083, but the larger change is in communication cost, which rises from 261.0 to 4981.6 bytes/token, a $19.1\times$ increase. The system still avoids some ineffective waiting, but consulted steps now consume much larger remote payloads.
 
\textbf{Unified design.}
Neither component alone recovers the full benefit. Relative to DRAGON, CONCORD increases throughput from 0.765 to 1.286 QPS while reducing communication from 256517.7 to 261.0 bytes/token. The two partial variants each recover only one side of this gain. NoSystem keeps communication small but leaves too much waiting on the critical path; NoComm retains some time-domain sparsity but loses the minimal-evidence advantage. The best operating point comes from combining both mechanisms.
 
\FloatBarrier
\subsection{Other Analyses}
\begin{figure}[!t]
\centering
\includegraphics[width=\columnwidth]{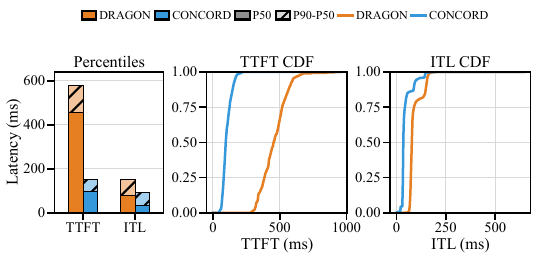}
\caption{Latency comparison between DRAGON and CONCORD. Left: TTFT and ITL at P50/P90 percentiles. Middle: TTFT cumulative distribution. Right: ITL cumulative distribution.}
\label{fig:latency_analysis}
\end{figure}

\textbf{Case study.}
Beyond aggregate throughput, CONCORD also reshapes how individual remote interactions happen. Figure~\ref{fig:latency_analysis} summarizes this shift from both percentile and distributional perspectives. In the left panel, DRAGON exhibits TTFT P50/P90 of 456.6/576.7\,ms versus 96.1/152.2\,ms for CONCORD, while ITL P50/P90 drops from 78.5/151.6\,ms to 35.2/92.0\,ms. The middle and right panels show the same trend in CDF form.
 
On the communication side, average messages per answer slightly increase from 23.24 to 28.77, yet bytes per token collapse from $2.04\times 10^5$ to $2.70\times 10^2$. Large coarse transfers are replaced by smaller ambiguity-targeted requests, consistent with certificate-guided supplementation.
 
\begin{figure}[!t]
\centering
\includegraphics[width=0.8\columnwidth]{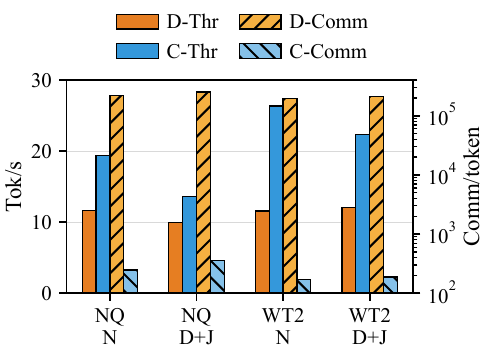}
\caption{Two-machine deployment results. DRAGON and CONCORD are compared under Native and Delay+Jitter link conditions on both NQ and WikiText-2, measured by throughput (tok/s) and communication cost (bytes/token).}
\label{fig:real_two_machine}
\end{figure}

\textbf{Deployment robustness.}
We further validate the trend under a real two-machine deployment. The cloud machine uses an Intel Xeon Platinum 8352V CPU with an RTX 4090D GPU, while the device machine uses an Intel Core i9-10900K CPU with an RTX 2070 SUPER GPU. Although the device machine is more powerful than a typical mobile device, the setup creates meaningful compute asymmetry (the device GPU has roughly one-third the throughput of the cloud GPU), and the gains stem from reduced waiting and communication rather than absolute hardware speed. Two link conditions are tested: direct LAN (\emph{Native}) and the same link with injected delay and jitter (\emph{Delay+Jitter}). As Figure~\ref{fig:real_two_machine} shows, the relative advantage of CONCORD persists across both tasks and both conditions. On NQ, CONCORD improves throughput by $1.66\times$ (Native) and $1.36\times$ (Delay+Jitter); on WikiText-2, the corresponding gains are $2.27\times$ and $1.85\times$. Communication drops from roughly 198k--252k bytes/token for DRAGON to 168.6--353.5 bytes/token for CONCORD across all four settings.
 
\textbf{Statistical analysis.}
To quantify the exactness boundary, Table~\ref{tab:protocol_diagnostics} reports consultation-path diagnostics from an instrumented 500-example, 3-seed run. About $21.66\pm1.55\%$ of generated tokens are committed locally without consultation, confirming that CONCORD does not require remote participation at every step. Among consulted steps, the remote side is often consequential: the final decision differs from the local top-1 on $22.29\pm1.33\%$ of them. Certificate success dominates ($99.98\pm0.03\%$) and forced fallback is almost absent ($0.017\pm0.030\%$); the residual fraction corresponds to steps still in progress at sequence termination. These numbers confirm the intended role separation: waiting control decides when consultation is worthwhile, while the communication layer remains effectively exact once consultation begins.
 
\begin{table}[!t]
\caption{Consultation-path diagnostics.}
\label{tab:protocol_diagnostics}
\centering
\small
\begin{tabular}{p{0.6\columnwidth}c}
\toprule
Metric & Value \\
\midrule
Consulted-step rate & $78.34\pm1.55\%$ \\
Local-only commit rate & $21.66\pm1.55\%$ \\
Local top-1 change among consulted steps & $22.29\pm1.33\%$ \\
Certificate success among consulted steps & $99.98\pm0.03\%$ \\
Forced fallback among consulted steps & $0.017\pm0.030\%$ \\
\bottomrule
\end{tabular}
\end{table}

\section{Conclusion}\label{sec:conclusion}
This paper presented CONCORD, a sparse aggregation framework for dual-end RAG under document isolation. By treating the cloud as an asynchronous evidence source, CONCORD couples waiting debt control with certificate-guided supplementation so that remote participation is reduced along both the temporal and communication dimensions. On consulted steps the greedy decision matches dense dual-end aggregation exactly, while steps that time out commit locally without remote evidence. Experiments on Natural Questions and WikiText-2 showed that CONCORD preserved answer quality and perplexity, improved throughput over DRAGON by about 66\% and 115\%, and reduced per-token communication by about 99.9\%. The current study is limited to greedy decoding in a two-end setting with a single model. For sampling-based decoding, the certificate mechanism can be extended to a probabilistic test that bounds the distance between the sparse and dense output distributions. For multi-node topologies, the debt controller can operate independently on each device-cloud link, since it relies only on local waiting outcomes. More broadly, the asynchronous sparse participation model may apply to other collaborative inference services where partial remote evidence arrives with variable delay.

\bibliographystyle{IEEEtran}
\bibliography{references}

\end{document}